\documentclass[11pt,a4paper]{article}
\usepackage[hyperref]{emnlp2018}
\usepackage{times}
\usepackage{latexsym}
\usepackage{color}
\usepackage{graphicx}
\usepackage{multirow}

\usepackage{url}
\usepackage{kky}

\aclfinalcopy 

\title{Contextual Encoding for Translation Quality Estimation}

\author{Junjie Hu, Wei-Cheng Chang, Yuexin Wu, Graham Neubig \\
  Language Technologies Institute, Carnegie Mellon University \\
  {\tt \{junjieh, wchang2, yuexinw, gneubig\}@cs.cmu.edu} }

\date{}

\begin{document}
\maketitle
\begin{abstract}
  The task of word-level quality estimation (QE) consists of taking a source sentence and machine-generated translation, and predicting which words in the output are correct and which are wrong.
  In this paper, propose a method to effectively encode the local and global contextual information for each target word using a three-part neural network approach.
  The first part uses an embedding layer to represent words and their part-of-speech tags in both languages. The second part leverages a one-dimensional convolution layer to integrate local context information for each target word. The third part applies a stack of feed-forward and recurrent neural networks to further encode the global context in the sentence before making the predictions. This model was submitted as the CMU entry to the WMT2018 shared task on QE, and achieves strong results, ranking first in three of the six tracks.%
\footnote{Our software is available at \url{https://github.com/junjiehu/CEQE}. } 
\end{abstract}

\section{Introduction}

Quality estimation (QE) refers to the task of measuring the quality of machine translation (MT) system outputs without reference to the gold translations \cite{blatz2004confidence,specia2013quest}.
QE research has grown increasingly popular due to the improved quality of MT systems, and potential for reductions in post-editing time and the corresponding savings in labor costs \cite{specia2011exploiting,turchi2014adaptive}.
QE can be performed on multiple granularities, including at word level, sentence level, or document level.
In this paper, we focus on quality estimation at word level, which is framed as the task of performing binary classification of translated tokens, assigning ``OK'' or ``BAD'' labels. 

Early work on this problem mainly focused on hand-crafted features with simple regression/classification models \cite{ueffing2007word,biccici2013referential}.
Recent papers have demonstrated that utilizing recurrent neural networks (RNN) can result in large gains in QE performance \cite{TACL1113}.
However, these approaches encode the context of the target word by merely concatenating its left and right context words, giving them limited ability to control the interaction between the local context and the target word.

In this paper, we propose a neural architecture, Context Encoding Quality Estimation (CEQE), for better encoding of context in word-level QE.
Specifically, we leverage the power of both (1) convolution modules that automatically learn local patterns of surrounding words, and (2) hand-crafted features that allow the model to make more robust predictions in the face of a paucity of labeled data.
Moreover, we further utilize stacked recurrent neural networks to capture the long-term dependencies and global context information from the whole sentence.

We tested our model on the official benchmark of the WMT18 word-level QE task.
On this task, it achieved highly competitive results, with the best performance over other competitors
on English-Czech, English-Latvian (NMT) and English-Latvian (SMT) word-level QE task, and ranking second place on English-German (NMT) and German-English word-level QE task.

\section{Model}

\begin{figure*}
	\centering
    \includegraphics[scale=0.35]{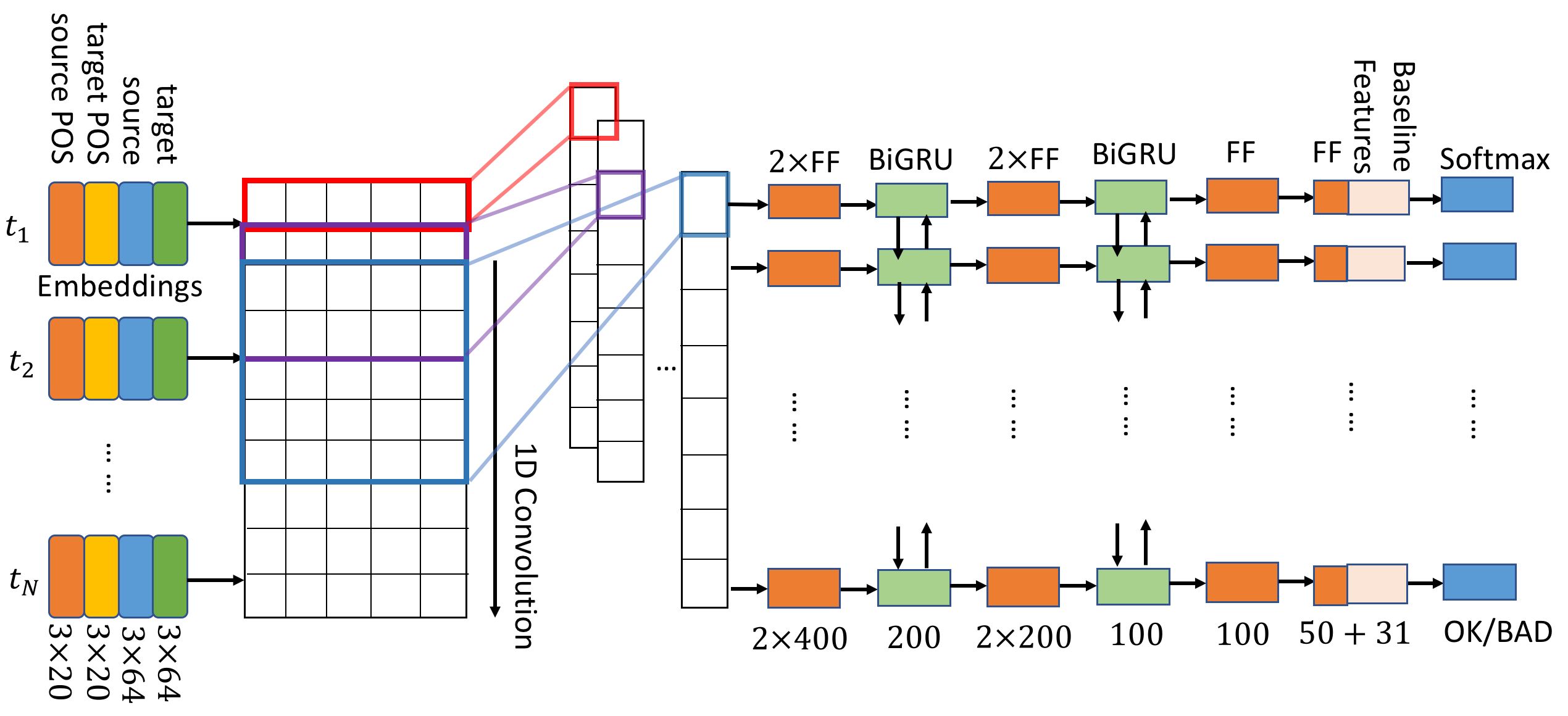}
    	\caption{The architecture of our model, with the convolutional encoder on the left, and stacked RNN on the right.}
	\label{fig:model}
\end{figure*}

The QE module receives as input a tuple $\langle s, t, \Acal \rangle$,
where $s = s_1, \ldots, s_M$ is the source sentence,
$t = t_1, \ldots, t_N$ is the translated sentence,
and $\Acal \subseteq \{ (m, n) | 1 \leq m \leq M, 1 \leq n \leq N \}$
is a set of word alignments.
It predicts as output a sequence $\hat{y} = y_1, \ldots, y_N$,
with each $y_i \in \{\text{BAD}, \text{OK}\}$.
The overall architecture is shown in Figure \ref{fig:model}

CEQE consists of three major components: (1) embedding layers for words and part-of-speech (POS) tags in both languages, (2) convolution encoding of the local context for each target word, and (3) encoding the global context by the recurrent neural network.    

\subsection{Embedding Layer}

Inspired by~\cite{TACL1113}, the first embedding layer is a vector representing each target word $t_j$ obtained by concatenating
the embedding of that word with those of the aligned words $s_{\Acal(:,t_j)}$ in the source. If a target word is aligned to multiple source words, we average the embedding of all the source words, and concatenate the target word embedding with its average source embedding. The immediate left and right contexts for source and target words are also concatenated,
enriching the local context information of the embedding of target word $t_j$.
Thus, the embedding of target word $t_j$, denoted as $\xb_j$, is a $6d$ dimensional vector,
where $d$ is the dimension of the word embeddings.
The source and target words use the same embedding parameters, and thus identical words in both languages, such as digits and proper nouns, have the same embedding vectors.
This allows the model to easily identify identical words in both languages. Similarly, the POS tags in both languages share the same embedding parameters. Table~\ref{tab:pos}
shows the statistics of the set of POS tags over all language pairs. 

\begin{table}[htp]
\setlength\tabcolsep{3pt} 
\centering
\begin{tabular}{lll} \hline
Language Pairs & Source & Target \\\hline
En-De (SMT) & 50 & 57   \\
En-De (NMT) & 49 & 58  \\
De-En & 58 & 50   \\
En-Lv (SMT) & 140 & 38   \\
En-Lv (NMT) & 167 & 43  \\
En-Cz & 440 & 57  \\\hline
\end{tabular}
\caption{Statistics of POS tags over all language pairs \label{tab:pos}}
\end{table}

\subsection{One-dimensional Convolution Layer}

The main difference between the our work and the neural model of \citet{TACL1113} is the
one-dimensional convolution layer.
Convolutions provide a powerful way to extract local context features, analogous to implicitly learning $n$-gram features. We now describe this integral part of our model.

After embedding each word in the target sentence $\{t_1, \ldots, t_j, \ldots, t_N\}$,
we obtain a matrix of embeddings for the target sequence,
\begin{equation*}
	\xb_{1:N} = \xb_1 \oplus \xb_2 \ldots \oplus \xb_N,
\end{equation*}
where $\oplus$ is the column-wise concatenation operator.
We then apply one-dimensional convolution \cite{kim2014convolutional,liu2017deep}
on $\xb_{1:N}$ along the target sequence to extract the local context of each target word.
Specifically, a one-dimensional convolution involves a filter $\wb \in \RR^{hk}$, which is applied
to a window of $h$ words in target sequence to produce new features. 
\begin{equation*}
	c_i = f(\wb \cdot \xb_{i:i+h-1} + b),
\end{equation*}
where $b \in \RR$ is a bias term and $f$ is some functions. This filter is applied to each
possible window of words in the embedding of target sentence
$\{\xb_{1:h}, \xb_{2:h+1}, \ldots, \xb_{N-h+1:N} \}$ to produce features
\begin{equation*}
	\cbb = [c_1, c_2, \ldots, c_{N-h+1}].
\end{equation*}
By the padding proportionally to the filter size $h$
at the beginning and the end of target sentence, we can obtain
new features $\cbb_{pad} \in \RR^{N}$ of target sequence with output size equals to input sentence length $N$.
To capture various granularities of local context, we consider filters with multiple window sizes $\Hcal=\{1, 3, 5, 7\}$,
and multiple filters $n_f=64$ are learned for each window size.

The output of the one-dimensional convolution layer, $C \in \RR^{N \times |\Hcal| \cdot n_f}$,
is then concatenated with the embedding of POS tags of the target words, as well as its aligned source words,
to provide a more direct signal to the following recurrent layers.


\subsection{RNN-based Encoding}
After we obtain the representation of the source-target word pair by the convolution layer, we follow a similar architecture as~\cite{TACL1113} to refine the representation of the word pairs using feed-forward and recurrent networks.
\begin{enumerate}
\item Two feed-forward layers of size 400 with rectified linear units (ReLU; \citet{nair2010rectified});
\item One bi-directional gated recurrent unit (BiGRU; \citet{cho-EtAl:2014:EMNLP2014}) layer with hidden size 200, where the forward and backward hidden states are concatenated and further normalized by layer normalization~\cite{ba2016layer}.
\item Two feed-forward layers of hidden size 200 with rectified linear units;
\item One BiGRU layer with hidden size 100 using the same configuration of the previous BiGRU layer;
\item Two feed-forward layers of size 100 and 50 respectively with ReLU activation.
\end{enumerate}
We concatenate the 31 baseline features extracted by the Marmot\footnote{https://github.com/qe-team/marmot} toolkit with the last 50 feed-forward hidden features. The baseline features are listed in Table~\ref{tab:features}.  We then apply a softmax layer on the combined features to predict the binary labels.
\begin{table}
\centering
\small
\begin{tabular}{l||l} \hline
Category & Description \\  \hline
Binary & target word is a stopword \\
Binary & target word is a punctuation mark \\
Binary & target word is a proper noun \\
Binary & target word is a digit \\
\multirow{2}[0]{*}{Float} & backoff behavior of ngram $w_{i-2}$ $w_{i-1}$ $w_{i}$ \\ & ($w_{i}$ is the target word)\\
Float & backoff behavior of ngram $w_{i-1}$ $w_{i}$ $w_{i+1}$ \\
Float & backoff behavior of ngram $w_{i}$ $w_{i+1}$ $w_{i+2}$ \\
\multirow{2}[0]{*}{One-hot} & highest order of ngram that includes \\ & target word and its left context\\
\multirow{2}[0]{*}{One-hot} & highest order of ngram that includes \\ & target word and its right context\\
\multirow{2}[0]{*}{One-hot} & highest order of ngram that includes \\ & source word and its left context \\
\multirow{2}[0]{*}{One-hot} & highest order of ngram that includes \\ & source word and its right context \\ \hline
\end{tabular}
\caption{Baseline Features \label{tab:features}}
\end{table}

\section{Training}
We minimize the binary cross-entropy loss between the predicted outputs and the targets. We train our neural model with mini-batch size $8$ using Adam~\cite{kingma2014adam} with learning rate $0.001$ and decay the learning rate by multiplying $0.75$ if the F1-Multi score on the validation set decreases during the validation. Gradient norms are clipped within $5$ to prevent gradient explosion for feed-forward networks or recurrent neural networks. Since the training corpus is rather small, we use dropout \cite{srivastava2014dropout} with probability $0.3$ to prevent overfitting.

\section{Experiment}
We evaluate our CEQE model on the WMT2018 Quality Estimation Shared Task\footnote{http://statmt.org/wmt18/quality-estimation-task.html} for word-level English-German, German-English, English-Czech, and English-Latvian QE. Words in all languages are lowercased. The evaluation metric is the multiplication of F1-scores for the ``OK'' and ``BAD'' classes against the true labels. F1-score is the harmonic mean of precision and recall. In Table~\ref{tab:leaderboard}, our model achieves the best performance on three out of six test sets in the WMT 2018 word-level QE shared task.

\begin{table}[htp]
\setlength\tabcolsep{3pt} 
\begin{tabular}{lllll} \hline
Language Pairs & F1-BAD & F1-OK & F1-Multi & Rank\\\hline
En-De (SMT) & 0.5075 & 0.8394 & 0.4260 & 3\\
En-De (NMT) & 0.3565 & 0.8827 & 0.3147 & 2\\
De-En & 0.4906 & 0.8640 & 0.4239 & 2\\
En-Lv (SMT) & 0.4211 & 0.8592 & 0.3618 & 1 \\
En-Lv (NMT) & 0.5192 & 0.8268 & 0.4293 & 1\\
En-Cz & 0.5882 & 0.8061 & 0.4741 & 1 \\\hline
\end{tabular}
\caption{Best performance of our model on six datasets in the WMT2018 word-level QE shared task on the leader board (updated on July 27th 2018) \label{tab:leaderboard}}
\end{table}

\subsection{Ablation Analysis}

In Table~\ref{tab:ablation}, we show the ablation study of the features used in our model on English-German, German-English, and English-Czech. For each language pair, we show the performance of CEQE without adding the corresponding components specified in the second column respectively. The last row shows the performance of the complete CEQE with all the components. As the baseline features released in the WMT2018 QE Shared Task for English-Latvian are incomplete, we train our CEQE model without using such features. We can glean several observations from this data:
\begin{enumerate}
	\item Because the number of ``OK'' tags is much larger than the number of ``BAD'' tags, the model is easily biased towards predicting the ``OK'' tag for each target word. The F1-OK scores are higher than the F1-BAD scores across all the language pairs.
    \item For German-English, English Czech, and English-German (SMT), adding the baseline features can significantly improve the F1-BAD scores.  
    \item For English-Czech, English-German (SMT), and English-German (NMT), removing POS tags makes the model more biased towards predicting ``OK'' tags, which leads to higher F1-OK scores and lower F1-BAD scores.  
    \item Adding the convolution layer helps to boost the performance of F1-Multi, especially on English-Czech and English-Germen (SMT) tasks. Comparing the F1-OK scores of the model with and without the convolution layer, we find that adding the convolution layer help to boost the F1-OK scores when translating from English to other languages, i.e., English-Czech, English-German (SMT and NMT). We conjecture that the convolution layer can capture the local information more effectively from the aligned source words in English. 
\end{enumerate}

\begin{table*}[htp]
\setlength\tabcolsep{3pt}
\centering
\begin{tabular}{lllll}
\hline
Language Pairs & Method & F1-BAD & F1-OK & F1-Multi \\ \hline
\multirow{5}[0]{*}{De-En} & - (Convolution + POS + features) & 0.4774 &  0.8680 &	0.4144 \\
& - (POS + features) & 0.4948 & 0.8474 & 0.4193 \\
& - features & 0.5095 & \bf{0.8735} & 0.4450 \\
& - POS & 0.4906 & 0.8640 & 0.4239  \\ 
& CEQE & \bf{0.5233} & 0.8721 & \bf{0.4564} \\ \hline
\multirow{5}[0]{*}{En-Cz} & - (Convolution + POS + features) & 0.5748 &	0.7622 &	0.4381 \\
& - (POS + features) & 0.5628 & 0.8000 & 0.4502 \\
& - features & 0.5777 & 0.7997 & 0.4620 \\
& - POS & 0.5192 & \bf{0.8268} & 0.4293 \\ 
& CEQE & \bf{0.5884} & 0.7991 & \bf{0.4702} \\ \hline
\multirow{5}[0]{*}{En-De (SMT)} & - (Convolution + POS + features) & 0.4677 &	0.8038 &	0.3759 \\
& - (POS + features) & 0.4768 & 0.8166 & 0.3894 \\
& - features & 0.4902 & 0.8230 & 0.4034 \\
& - POS & 0.5047 & \bf{0.8431} & 0.4255 \\
& CEQE & \bf{0.5075} & 0.8394 & \bf{0.4260} \\ \hline
\multirow{5}[0]{*}{En-De (NMT)} & - (Convolution + POS + features) & 0.3545 & 0.8396 &	0.2976 \\
& - (POS + features) & 0.3404 & 0.8752 & 0.2979 \\  
& - features & \bf{0.3565} & 0.8827 & \bf{0.3147} \\
& - POS & 0.3476 & \bf{0.8948} & 0.3111 \\ 
& CEQE & 0.3481 & 0.8835 & 0.3075 \\ \hline
\end{tabular}
\caption{Ablation study on the WMT18 Test Set \label{tab:ablation}}
\end{table*}

\section{Case Study}
Table~\ref{tab:example} shows two examples of quality prediction on the validation data of WMT2018 QE task for English-Czech. In the first example, the model without POS tags and baseline features is biased towards predicting ``OK'' tags, while the model with full features can detect the reordering error. In the second example, the target word ``panelu'' is a variant of the reference word ``panel''.  The target word ``znaky'' is the plural noun of the reference ``znak''. Thus, their POS tags have some subtle differences. Note the target word ``zmnit'' and its aligned source word ``change'' are both verbs. We can observe that POS tags can help the model capture such syntactic variants.   

\begin{table*}[tbh]
\setlength\tabcolsep{3pt} 
\centering
\begin{tabular}{ll}
\hline
Source & specify the scope of blending options : \\
MT & ur\v{c}ete rozsah {\color{blue}prolnut\'{i} voleb} :  \\
Reference & ur\v{c}ete rozsah {\color{blue}voleb prolnut\'{i}} :  \\
no POS \& features & {\color{green}ur\v{c}ete rozsah prolnut\'{i} voleb :}  \\
CEQE & {\color{green}ur\v{c}ete rozsah {\color{red}prolnut\'{i} voleb} :}  \\ \hline 

Source & use the Character panel and Paragraphs panel to change the appearance of text . \\
MT & {\color{blue}pomoc\'{i} panelu znaky} a odstavce , chcete - li zm\v{e}nit vzhled textu .\\
Reference & {\color{blue}pou\v{z}ijte panel znak} a panel odstavce , chcete - li zm\v{e}nit vzhled textu .\\
no POS \& features & {\color{red}pomoc\'{i}} {\color{green}panelu znaky a odstavce} {\color{red}, chcete - li změnit} {\color{green}vzhled textu .}\\
CEQE & {\color{red}pomoc\'{i} panelu znaky} {\color{green}a odstavce {\color{red}, chcete} - li zm\v{e}nit vzhled textu .}\\
\hline 
\end{tabular}
\caption{Examples on WMT2018 validation data. The source and translated sentences, the reference sentences, the predictions of the CEQE without and with POS tags and baseline features are shown. Words predicted as OK are shown in {\color{green}green}, those predicted as BAD are shown in {\color{red}red}, the difference between the translated and reference sentences are shown in {\color{blue}blue}.\label{tab:example}} 
\end{table*}

\subsection{Sensitivity Analysis}
\begin{figure}
	\centering
    \includegraphics[width=0.48\textwidth]{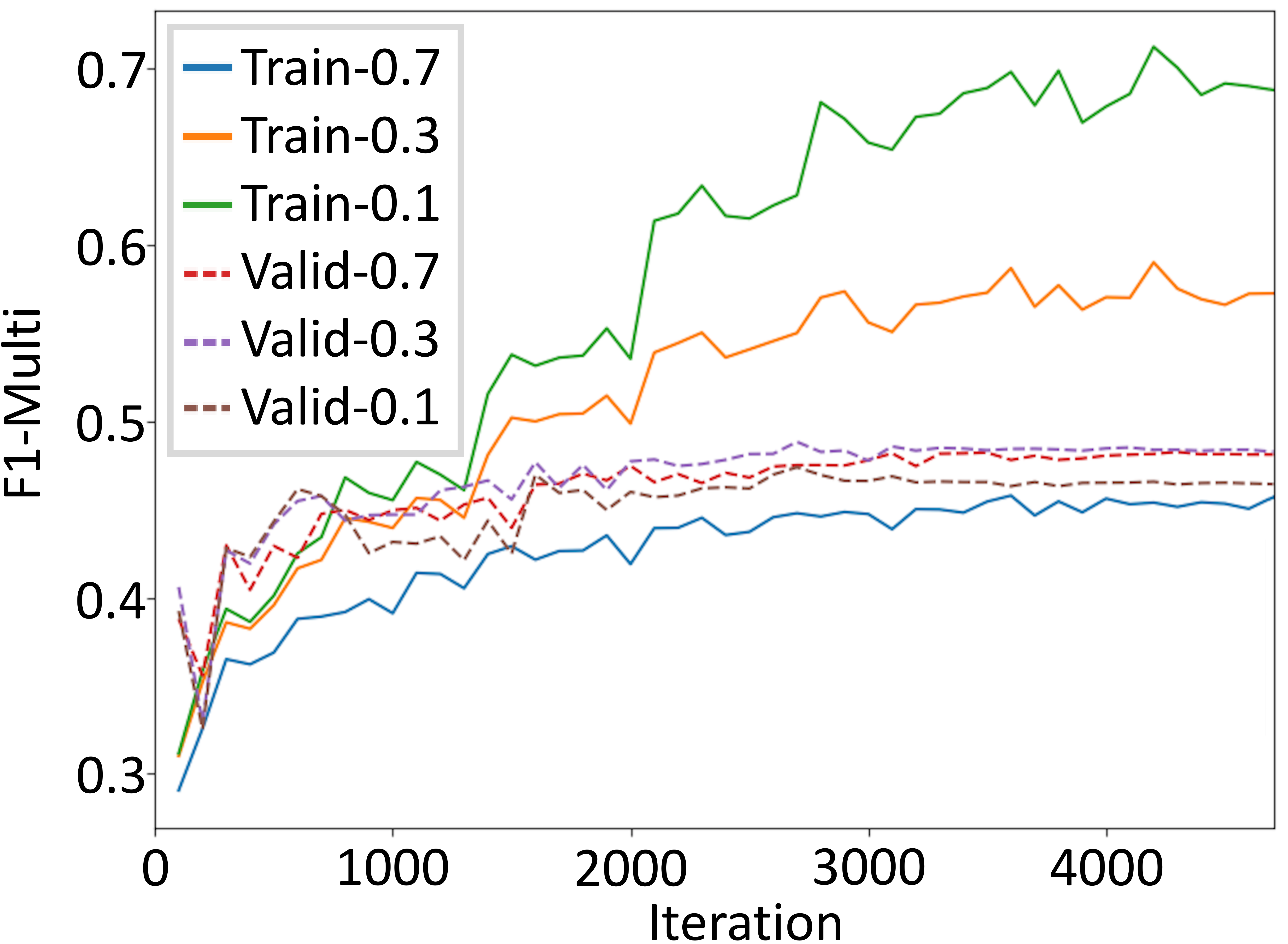}
	\caption{Effect of the dropout rate during training.} 
	\label{fig:dropout}
\end{figure}

During training, we find that the model can easily overfit the training data, which yields poor performance on the test and validation sets. To make the model more stable on the unseen data, we apply dropout to the word embeddings, POS embeddings, vectors after the convolutional layers and the stacked recurrent layers. In Figure~\ref{fig:dropout}, we examine the accuracies dropout rates in $[0.1, 0.3, 0.7]$. We find that adding dropout alleviates overfitting issues on the training set. If we reduce the dropout rate to $0.1$, which means randomly setting some values to zero with probability $0.1$, the training F1-Multi increases rapidly and the validation F1-multi score is the lowest among all the settings. Preliminary results proved best for a dropout rate of $0.3$, so we use this in all the experiments.   

\section{Conclusion}
In this paper, we propose a deep neural architecture for word-level QE.
Our framework leverages a one-dimensional convolution on the
concatenated word embeddings of target and its aligned source words to extract salient local feature maps.
In additions, bidirectional RNNs are applied to capture temporal dependencies for better sequence prediction.
We conduct thorough experiments on four language pairs in the WMT2018 shared task.
The proposed framework achieves highly competitive results, outperforms all other participants on English-Czech and English-Latvian word-level, and is
second place on English-German, and German-English language pairs.  

\section*{Acknowledgements}

The authors thank Andre Martins for his advice regarding the word-level QE task.

This work is sponsored by Defense Advanced Research Projects Agency Information Innovation Office (I2O). Program: Low Resource Languages for Emergent Incidents (LORELEI). Issued by DARPA/I2O under Contract No. HR0011-15-C0114. The views and conclusions contained in this document are those of the authors and should not be interpreted as representing the official policies, either expressed or implied, of the U.S. Government. The U.S. Government is authorized to reproduce and distribute reprints for Government purposes notwithstanding any copyright notation here on.

\bibliography{emnlp2018.bib}

\begin{thebibliography}{14}
\expandafter\ifx\csname natexlab\endcsname\relax\def\natexlab#1{#1}\fi

\bibitem[{Ba et~al.(2016)Ba, Kiros, and Hinton}]{ba2016layer}
Lei~Jimmy Ba, Ryan Kiros, and Geoffrey~E. Hinton. 2016.
\newblock Layer normalization.
\newblock \emph{CoRR}, abs/1607.06450.

\bibitem[{Bi{\c{c}}ici(2013)}]{biccici2013referential}
Ergun Bi{\c{c}}ici. 2013.
\newblock Referential translation machines for quality estimation.
\newblock In \emph{Proceedings of the eighth workshop on statistical machine
  translation}, pages 343--351.

\bibitem[{Blatz et~al.(2004)Blatz, Fitzgerald, Foster, Gandrabur, Goutte,
  Kulesza, Sanchis, and Ueffing}]{blatz2004confidence}
John Blatz, Erin Fitzgerald, George Foster, Simona Gandrabur, Cyril Goutte,
  Alex Kulesza, Alberto Sanchis, and Nicola Ueffing. 2004.
\newblock Confidence estimation for machine translation.
\newblock In \emph{Proceedings of the 20th international conference on
  Computational Linguistics}, page 315. Association for Computational
  Linguistics.

\bibitem[{Cho et~al.(2014)Cho, van Merrienboer, Gulcehre, Bahdanau, Bougares,
  Schwenk, and Bengio}]{cho-EtAl:2014:EMNLP2014}
Kyunghyun Cho, Bart van Merrienboer, Caglar Gulcehre, Dzmitry Bahdanau, Fethi
  Bougares, Holger Schwenk, and Yoshua Bengio. 2014.
\newblock Learning phrase representations using rnn encoder--decoder for
  statistical machine translation.
\newblock In \emph{Proceedings of the 2014 Conference on Empirical Methods in
  Natural Language Processing (EMNLP)}, pages 1724--1734, Doha, Qatar.
  Association for Computational Linguistics.

\bibitem[{Kim(2014)}]{kim2014convolutional}
Yoon Kim. 2014.
\newblock Convolutional neural networks for sentence classification.
\newblock In \emph{Proceedings of the 2014 Conference on Empirical Methods in
  Natural Language Processing, {EMNLP} 2014, October 25-29, 2014, Doha, Qatar,
  {A} meeting of SIGDAT, a Special Interest Group of the {ACL}}, pages
  1746--1751.

\bibitem[{Kingma and Ba(2015)}]{kingma2014adam}
Diederik~P Kingma and Jimmy Ba. 2015.
\newblock Adam: A method for stochastic optimization.
\newblock In \emph{ICLR}.

\bibitem[{Liu et~al.(2017)Liu, Chang, Wu, and Yang}]{liu2017deep}
Jingzhou Liu, Wei-Cheng Chang, Yuexin Wu, and Yiming Yang. 2017.
\newblock Deep learning for extreme multi-label text classification.
\newblock In \emph{Proceedings of the 40th International ACM SIGIR Conference
  on Research and Development in Information Retrieval}, pages 115--124. ACM.

\bibitem[{Martins et~al.(2017)Martins, Junczys-Dowmunt, Kepler, Astudillo,
  Hokamp, and Grundkiewicz}]{TACL1113}
Andr{\'e} Martins, Marcin Junczys-Dowmunt, Fabio Kepler, Ram{\'o}n Astudillo,
  Chris Hokamp, and Roman Grundkiewicz. 2017.
\newblock Pushing the limits of translation quality estimation.
\newblock \emph{Transactions of the Association for Computational Linguistics},
  5:205--218.

\bibitem[{Nair and Hinton(2010)}]{nair2010rectified}
Vinod Nair and Geoffrey~E Hinton. 2010.
\newblock Rectified linear units improve restricted boltzmann machines.
\newblock In \emph{Proceedings of the 27th international conference on machine
  learning (ICML-10)}, pages 807--814.

\bibitem[{Specia(2011)}]{specia2011exploiting}
Lucia Specia. 2011.
\newblock Exploiting objective annotations for measuring translation
  post-editing effort.
\newblock In \emph{Proceedings of the 15th Conference of the European
  Association for Machine Translation}, pages 73--80.

\bibitem[{Specia et~al.(2013)Specia, Shah, Souza, and Cohn}]{specia2013quest}
Lucia Specia, Kashif Shah, Jose~GC Souza, and Trevor Cohn. 2013.
\newblock Quest-a translation quality estimation framework.
\newblock In \emph{Proceedings of the 51st Annual Meeting of the Association
  for Computational Linguistics: System Demonstrations}, pages 79--84.

\bibitem[{Srivastava et~al.(2014)Srivastava, Hinton, Krizhevsky, Sutskever, and
  Salakhutdinov}]{srivastava2014dropout}
Nitish Srivastava, Geoffrey Hinton, Alex Krizhevsky, Ilya Sutskever, and Ruslan
  Salakhutdinov. 2014.
\newblock Dropout: a simple way to prevent neural networks from overfitting.
\newblock \emph{The Journal of Machine Learning Research}, 15(1):1929--1958.

\bibitem[{Turchi et~al.(2014)Turchi, Anastasopoulos, de~Souza, and
  Negri}]{turchi2014adaptive}
Marco Turchi, Antonios Anastasopoulos, Jos{\'e}~GC de~Souza, and Matteo Negri.
  2014.
\newblock Adaptive quality estimation for machine translation.
\newblock In \emph{Proceedings of the 52nd Annual Meeting of the Association
  for Computational Linguistics (Volume 1: Long Papers)}, volume~1, pages
  710--720.

\bibitem[{Ueffing and Ney(2007)}]{ueffing2007word}
Nicola Ueffing and Hermann Ney. 2007.
\newblock Word-level confidence estimation for machine translation.
\newblock \emph{Computational Linguistics}, 33(1):9--40.

\end{thebibliography}
\bibliographystyle{acl_natbib_nourl}

\end{document}